\pdfoutput=1

\documentclass[11pt]{article}

\usepackage[]{acl}

\usepackage{times}
\usepackage{latexsym}
\usepackage{amsmath}
\usepackage{amsfonts}
\usepackage{amssymb}
\usepackage{enumitem}
\usepackage{subfig}
\usepackage{url}
\usepackage{hyperref}

\usepackage{inconsolata}
\usepackage{booktabs}
\usepackage{graphicx}
\graphicspath{{latex/}{./}}

\usepackage[T1]{fontenc}

\usepackage[utf8]{inputenc}

\usepackage[para]{footmisc}

\usepackage{microtype}

\newcommand{\RedwoodSize}{451}
\newcommand{\RedwoodDatasets}{13}
\newcommand{\RedwoodNaiveSize}{619}
\newcommand{\RedwoodQueries}{62,216}
\newcommand{\RedwoodNaiveQueries}{85,746}
\newcommand{\OOSTotalQueries}{3,267}
\newcommand{\MetaDatasetSize}{722}

\title{Redwood: Using Collision Detection to Grow \\a Large-Scale Intent Classification Dataset}

\author{Stefan Larson \and Kevin Leach \\
         Vanderbilt University\\
         \texttt{\{firstname.lastname\}@vanderbilt.edu}}

\begin{document}
\maketitle
\begin{abstract}
Dialog systems must be capable of incorporating new skills via updates over time in order to reflect new use cases or deployment scenarios.
Similarly, developers of such ML-driven systems need to be able to add new training data to an already-existing dataset to support these new skills.
In intent classification systems, problems can arise if training data for a new skill's intent overlaps semantically with an already-existing intent.
We call such cases \emph{collisions}.
This paper introduces the task of intent collision detection between multiple datasets
for the purposes of growing a system's skillset.
We introduce several methods for detecting collisions, and evaluate our methods on real datasets that exhibit collisions.
To highlight the need for intent collision detection, we show that model performance suffers if new data is added in such a way that does not arbitrate colliding intents.
Finally, we use collision detection to construct and benchmark a new dataset, \emph{Redwood}, which is composed of \RedwoodSize~intent categories from \RedwoodDatasets~original intent classification datasets, making it the largest publicly available intent classification benchmark.
\end{abstract}

\section{Introduction}\label{sec:introduction}
As task-oriented dialog systems like Alexa and Siri have become more and more pervasive, tools enabling developers to build custom dialog systems have followed suit.
Such tools---like Microsoft's Luis\footnote{~\texttt{\href{https://www.luis.ai/}{luis.ai}}}, Twilio's Autopilot\footnote{~\texttt{\href{https://www.twilio.com/autopilot}{twilio.com/autopilot}}}, Rasa\footnote{~\texttt{\href{https://rasa.com/}{rasa.com}}}, and Google's DialogFlow\footnote{~\texttt{\href{https://cloud.google.com/dialogflow}{google.com/dialogflow}}}---enable engineers and dialog designers to craft dialog systems composed of \emph{intents}, or core categories of competencies or skills in which the system is knowledgeable and to which the system can respond intelligently.
New intents may be added periodically to the dialog system as part of its development and maintenance cycle, or dialog system models may be combined together (e.g., \citet{Clarke2022OneAT}).

These phenomena may occur especially in real-world deployments, where datasets for dialog models may be developed, grown, and modified by large (and even disparate) teams over the span of a project's lifetime.
Furthermore, dialog system models and their corresponding training datasets are sometimes offered as-a-service or ``off-the-shelf'' to dialog system builders who might not be fully familiar with the breadth or scope of the pre-existing dataset or model. 
If the builder adds a new intent to the dataset that overlaps with an existing intent, then the re-trained model's performance can suffer. 
As such, there is a need for tools and algorithms to help detect when a new intent overlaps---that is, \emph{collides}---with an already-existing intent category.

In this paper, we introduce the challenge of \emph{intent collision detection}, and develop several algorithms for determining whether a candidate intent category collides with another intent category.
To do so, we curate and release a meta-dataset of \MetaDatasetSize~intents from \RedwoodDatasets~existing datasets.
This graph-like meta-dataset consists of annotations indicating tuples of colliding intent pairs (examples of colliding intents can be seen in Table~\ref{tab:example-collisions}).
We then introduce several collision detection algorithms and evaluate them on this meta-dataset. 

We also use intent collision detection to build \emph{Redwood}, a new intent classification dataset of \RedwoodSize~intent categories. \emph{Redwood} is built by combining \RedwoodDatasets~smaller datasets.
As a comparison, we also build \emph{Redwood-na\"{i}ve}, which is constructed by na\"i{v}ely joining together all \RedwoodDatasets~datasets without arbitrating colliding intents. 
We find that classifier performance on \emph{Redwood-na\"{i}ve} to be substantially worse than \emph{Redwood}, showcasing the negative effect of not addressing intent collisions in data.

\begin{table*}[]
    \centering\scalebox{0.675}{
    \begin{tabular}{cccc}
    \toprule
        \textbf{Dataset} &  & \textbf{Samples~~~~} & \\
        \midrule
       \emph{Snips}  &  \emph{how cold is it in princeton junction} & \emph{will it be chilly in fiji at ten pm} & \emph{is it foggy in shelter island}\\
       \emph{Clinc-150}  & \emph{give me the 7 day forecast} & \emph{what's the temperature like in tampa} & \emph{will it rain today} \\
       \emph{MTOP} & \emph{what is the weather in new york today} & \emph{how much is it going to rain tomorrow} & \emph{give me the weather for march 13th} \\
       \midrule
       \emph{Slurp} & \emph{set alarm tomorrow at 6 am} & \emph{make an alarm for 4pm} & \emph{set a wake up call for 10 am} \\
       \emph{MTOP} & \emph{can you set a warning alarm for 7pm} & \emph{set an alarm for monday at 5pm} & \emph{make an alarm for the 5th}\\
       \emph{Clinc-150} & \emph{wake me up at noon tomorrow} & \emph{set my alarm for getting up} & \emph{i need you to set alarm for me} \\
       \midrule
       \emph{HWU}  &  \emph{how much is 1gbp in usd} & \emph{what's the exchange rates} & \emph{how much is \$50 in pounds}\\
       \emph{Clinc-150}  & \emph{tell me five dollars in yen and rubles} & \emph{how many pesos in one dollar us} & \emph{usd to yen is what right now}\\
       \emph{Banking-77} & \emph{do you know the rate of exchange} & \emph{how is the exchange rate doing} & \emph{what are the current exchange rates} \\
       \midrule
       \emph{Clinc-150}  & \emph{please start calling me mandy} & \emph{I want you to call me this new name} & \emph{the name you should call me is janet}\\
       \emph{ACID} & \emph{how do i change my name} & \emph{need my name to be updated} & \emph{I need to fix my name in your system} \\
       \emph{Banking-77} & \emph{where can I find how to change my name} & \emph{details need to be modified} & \emph{after I got married I need to change my name}\\
       \midrule
       \emph{Snips}  &  \emph{play magic sam from the thirties} & \emph{play music by blowfly from the seventies} & \emph{play jeff pilson on youtube}\\
       \emph{DSTC-8}  & \emph{I want to hear the song high} & \emph{I would like to listen to touch it on tv} & \emph{I'd like to listen to the way I talk} \\
       \emph{HWU} & \emph{please play yesterday from beattles} & \emph{I'd like to hear queen's barcelona} & \emph{play daft punk} \\
       \midrule
       \emph{MetalWOz}  &  \emph{help me find restaurants in miami fl} & \emph{I need help finding a place to eat} & \emph{I need to find an italian restaurant in denver}\\
       \emph{DSTC-8}  & \emph{can you help find a place to eat} & \emph{I'm looking for a filipino place to eat} & \emph{I want to find a restaurant in albany}\\
       \emph{HWU} & \emph{find me a nice restaurant for dinner} & \emph{where can I get shawarma in this area} & \emph{what's the best chicken place near me} \\
       \midrule
       \emph{Outlier} & \emph{what is my balance} & \emph{update me on my account balance} & \emph{let me know how much money I have}\\
       \emph{Clinc-150} & \emph{what's my current checking balance} & \emph{what is the total of my bank accounts} & \emph{how much total cash do I have in the bank}\\
       \emph{DSTC-8} & \emph{I want to know my checking account balance} & \emph{I'd like to check my balance} & \emph{man how much money do I have in the bank} \\
       \bottomrule
    \end{tabular}}
    \caption{Examples of data that will trigger collisions. Each row of the table displays three samples from a single intent in a particular dataset.  Among these three samples, each line collides with an intent category from the other two datasets.
    }
    \label{tab:example-collisions}
\end{table*}

Upon official release, \emph{Redwood} will by far be the largest openly available intent classification dataset in terms of breadth of intent categories.
Our hope is that the new \emph{Redwood} dataset serves as a showcase for intent collision detection as well as a new, publicly-available, large-scale challenge dataset for intent classification models for dialog systems.
Both the collision meta-dataset and \emph{Redwood} are publicly available at \texttt{\href{https://github.com/gxlarson/redwood}{github.com/gxlarson/redwood}}.




\section{Related Work}

\paragraph{The Collision Detection Task.}
We discuss three areas of related work related to our proposed intent collision detection task: generalized zero-shot learning, open set classification, and out-of-domain (or out-of-scope) sample identification.

In generalized zero-shot learning (e.g., \citet{zhang-etal-2022-learn-gzs}), a model is trained with data from a set of ``seen'' label classes (e.g., intents) and, during inference, must identify test samples as belonging to either a ``seen'' label class or an ``unseen'' class for which the model has limited auxiliary knowledge (e.g., descriptions of unseen classes, but no concrete training examples).

Both open set classification and out-of-domain sample identification refer to the modeling task of 
classifying inference samples among label classes seen during training or to identify if the sample belongs to an unknown or undefined label class (e.g., \citet{chatbot150, zhang-etal-2021-textoir}).
Slot-filling models that are trained on B/I/O 
tags naturally predict the unknown class label as O tags, but for intent classifiers the task is much more challenging since it requires curating viable training data for an out-of-domain category (i.e., it is challenging to know in advance what types of out-of-domain inputs a system might encounter).

Our proposed task of intent collision detection differs from the aforementioned tasks because ``inference'' samples need not be considered one at a time, but can instead be grouped together into entire candidate intent categories. 
This enables considering entirely different modeling tasks
like those discussed in Section~\ref{sec:collision-detection-methods}.
Nevertheless, both our meta-dataset of intent collisions and \emph{Redwood} allow for the evaluation of both zero-shot and generalized zero-shot learning models, and the \emph{Redwood} intent classification dataset includes a substantial number of out-of-domain samples for evaluating open set classification and out-of-domain sample detection.

\paragraph{Intent Classification Corpora.}
There are several smaller corpora for evaluating intent classification models, some spanning broad domains (e.g., \citet{sds-eval}, \citet{chatbot150}, \citet{li2021mtop}) and others focusing fine-grained evaluation of individual domains (e.g., the Banking-77 corpus \cite{banking77} with respect to the personal banking domain).
While most datasets are constructed via crowdsourcing, our new Redwood dataset is constructed from both (1) already existing datasets and (2) newly crowdsourced intents.



\paragraph{Dataset Derivation and Combination.}
Datasets are sometimes formed from other datasets, either by deriving a new dataset from an existing one, or by combining datasets together.
The former category include translations of dialog datasets (e.g., \cite{multilingual-atis, multiatispp}) as well as re-formulations of existing datasets into new tasks (e.g., converting a semantic role labeling (SRL) dataset to open information extraction (OIE) data as done in \citet{solawetz-larson-2021-lsoie}).

Dataset combination has been used in other fields beyond dialog systems and conversational AI. For instance, \citet{song-etal-2020-adversarial} combined several speech recognition datasets together to form their \emph{SpeechStew} dataset. As there are no target labels analogous to intents in automatic speech recognition, the creators of \emph{SpeechStew} did not have to consider collisions among intent categories.
In this paper, our focus is primarily on dataset combination, but we also derive intent classification data from several turn-based dialog corpora (\emph{MetalWOz} and \emph{DSTC-8}, discussed in Section~\ref{sec:datasets}).


\section{Detecting Collisions}

In this section we discuss our proposed challenge,
intent collision detection. We begin with a motivating example showing why detecting collisions is important, as well as a formal problem statement. Then, we introduce and evaluate several collision detection baselines on our meta-dataset.

\begin{figure}
    \centering
    \includegraphics[clip,width=\columnwidth]{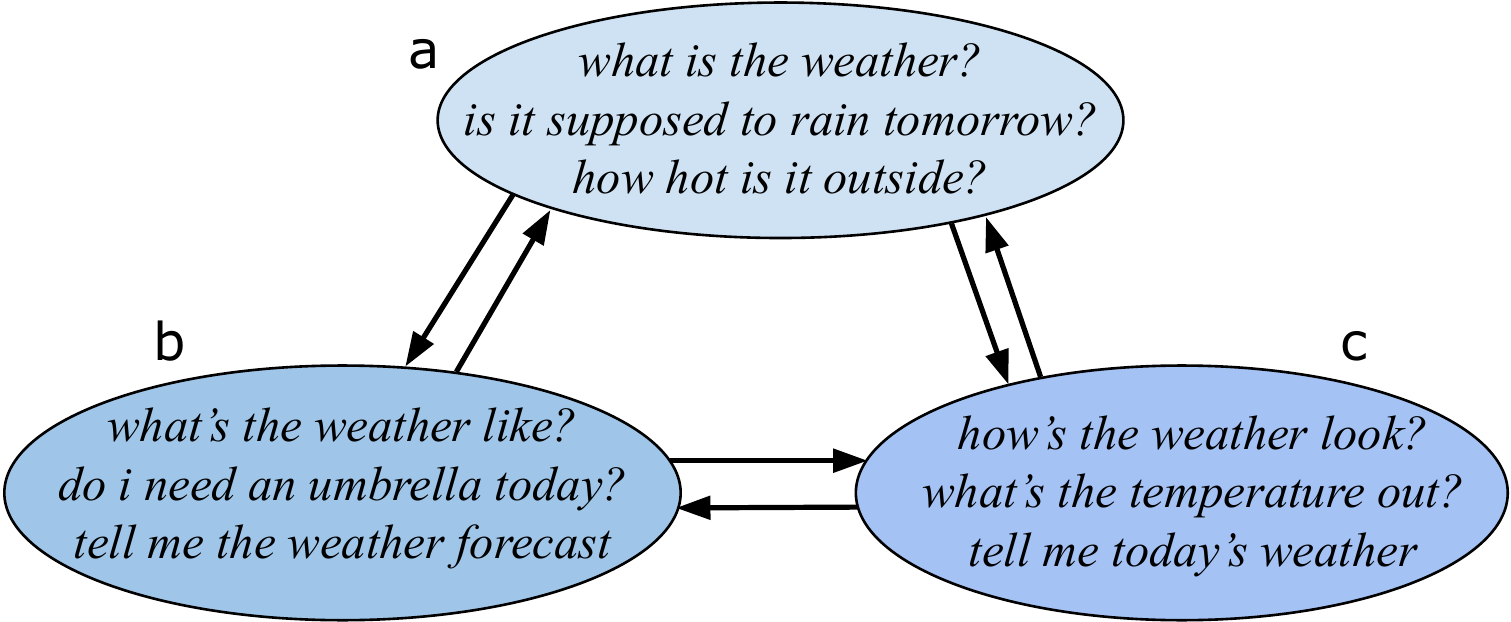}
    \caption{Transitive collisions.}
    \label{fig:transitive-collisions}
\end{figure}

\subsection{Motivating Example}\label{sec:motivating-example}

As a motivating example, suppose our intent classification system has been trained on the \emph{Clinc-150} dataset \cite{chatbot150}, an intent classification dataset consisting of 150 intents.\footnote{In this paper, dataset names are in \emph{italics} and intent names are in \texttt{teletype font}. Example queries are in \emph{italics} and in quotes if they appear in-line.}
The \emph{Clinc-150} dataset includes an intent called \texttt{weather}, which is meant to handle weather-related queries such as ``\emph{what's the weather like today}'' and ``\emph{tell me the weather in New York}.''
Suppose further that a new developer or a new team attempts to update the intent classifier with new data that contains a new intent category, such as the \texttt{get\_weather} intent from the \emph{HWU} dataset\footnote{Recall from Section~\ref{sec:introduction} that such updates from new teams or new developers may be from routine perfective maintenance during a model's lifetime.} \cite{sds-eval}.
In such a scenario, there are now training data samples that overlap substantially, but that are labeled with different intents (\texttt{weather} vs. \texttt{get\_weather} in this example).
Thus, upon updating the model by training on \emph{HWU}'s \texttt{get\_weather} data, the predictive performance on any weather-related inference queries might be split between these two intents.
This disparity can also cause unintended consequences downstream in production models,
such as calls to database systems that are triggered based on the user's intent. 

Indeed, when we train a BERT classifier on the original \emph{Clinc-150} training set, the accuracy on the \texttt{weather} test set is 100\%. 
When we add a \emph{HWU}'s \texttt{get\_weather} intent to \emph{Clinc-150} to create a new 151$^{st}$ intent and re-train the BERT classifier, we observe an accuracy score of 60\% on the \texttt{weather} test set.
This performance drop is a symptom of having added an intent category that collides with another intent category.
Such a model---which was trained on colliding intents---could cause unexpected behavior on downstream events, especially if the \texttt{weather} and \texttt{get\_weather} intents trigger different business logic workflows or system responses. 
We note that, while in this example, the colliding \texttt{weather} and \texttt{get\_weather} intent names are quite similar, other colliding pairs like Snips' \texttt{search\_screening\_event} and MetalWOz's \texttt{movie\_listings} do not have lexically similar intent names, precluding straightforward string matching of intent names.

\begin{figure}
    \centering
    \includegraphics[clip,width=\columnwidth]{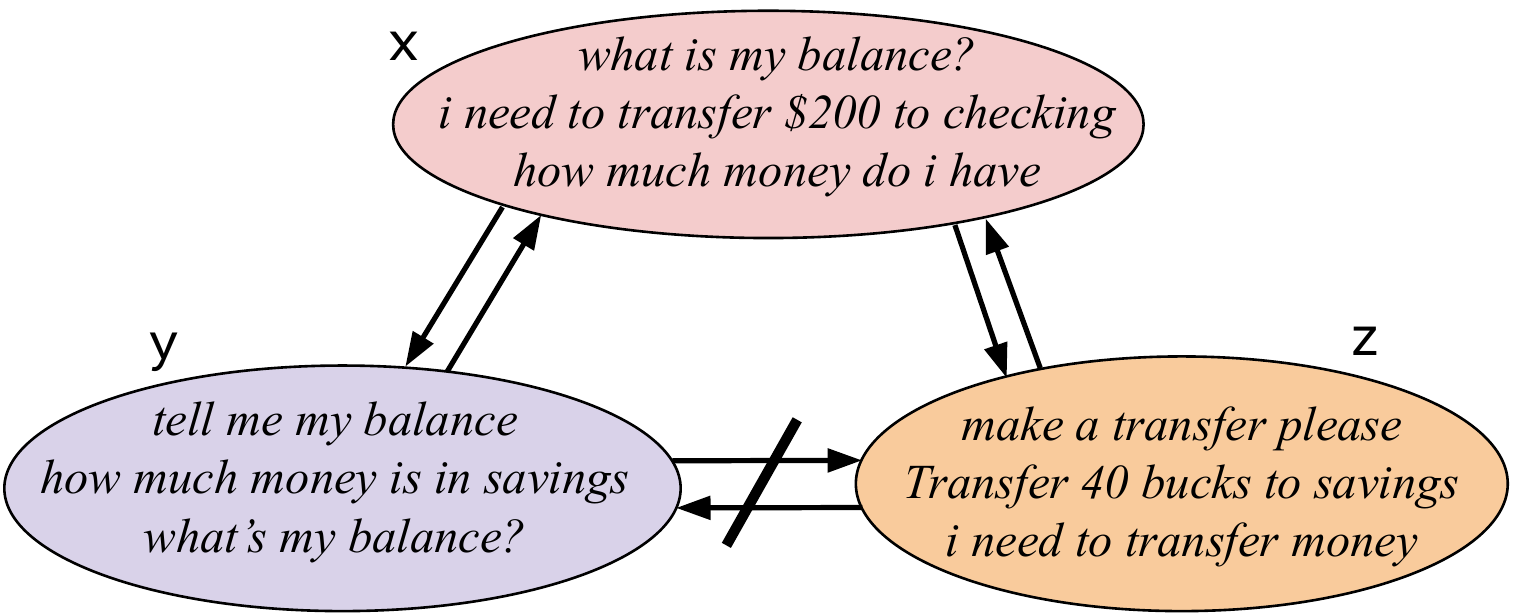}
    \caption{Non-transitive collisions.}
    \label{fig:nontransitive}
\end{figure}


\subsection{Problem Statement}\label{sec:problem-statement}


In this subsection, we formally define our collision detection problem. 
We first consider a scenario in which we have two intent classification datasets, $\mathcal{A}$ and $\mathcal{B}$, where $A_i \in \mathcal{A}$ and $B_j \in \mathcal{B}$ refer to specific intent categories in each. 
We say that intent categories $A_i$ and $B_j$ \emph{collide} if there exist a sufficient number of queries in $A_i$ that \emph{semantically overlap} with a sufficient number of queries in $B_j$.
This \emph{semantic overlap} can occur when a developer attempts to add new intent categories to a starting training dataset---when an intent classification model trained on the combined dataset $\mathcal{A} \cup \mathcal{B}$ will cause queries belonging to $A_i$ to be classified in $B_j$ (and vice versa).

As an example, suppose we have an intent classifier built from a starting dataset such as \emph{Clinc-150}, which, among other things, contains a \texttt{weather} intent category for weather-related inquiries (cf. Section~\ref{sec:motivating-example}). 
Suppose further that we seek to grow this starting dataset by adding datapoints from a candidate dataset such as \emph{HWU} (see Section~\ref{sec:motivating-example}, which contains a \texttt{get\_weather} intent category). 
If we na\"{i}vely combine these two datasets together, a resulting intent classifier will result in some queries from the original \texttt{weather} category to be classified to the newly-added \texttt{get\_weather} category because these two categories are semantically similar. 
Table~\ref{tab:example-collisions} illustrates several example colliding intents and associated queries.
Our approach addresses these collisions by detecting their prevalence and quantifying their impact automatically, aiding developers in improving the quality of their datasets and scope of their dialogue systems.

Because the notion of semantic overlap can differ from category to category and dataset to dataset, 
we observe several classes of relationships among colliding intent categories in practice. 
In particular, intent collisions can be \emph{simple-pairwise}, \emph{transitive}, or \emph{hierarchical}. 
In the simple-pairwise case, two intents collide with each other \emph{only}, and not with any other intent in either dataset.
However, we also observe transitivity within intent classes.  Figure~\ref{fig:transitive-collisions} illustrates example utterances within intent classes \texttt{a}, \texttt{b}, and \texttt{c}, where all intent classes are transitively related to one another in a cycle. 

Lastly, we observe non-transitive hierarchies among colliding intents. 
In this case, a broad intent category from one dataset can collide with two or more intent categories that do not relate to each other.
Figure~\ref{fig:nontransitive} shows a hypothetical intent class \texttt{x} consisting of general banking queries, including balance inquiries and transfer requests, and classes \texttt{y} and \texttt{z} consist solely of balance inquiry and transfer requests, respectively. 
Here, because class \texttt{x} is more broad than \texttt{y} and \texttt{z}, each of \texttt{y} and \texttt{z} collide with \texttt{x}, but \texttt{y} and \texttt{z} do not collide with each other. 
Our approach can help developers reveal such cases when managing datasets, and we consider these collision relationships in the creation of our Redwood dataset.

\subsection{Approaches}\label{sec:collision-detection-methods}

We introduce two approaches for detecting collisions: \emph{Classifier Confusion} and \emph{Data Coverage}.

\paragraph{Classifier Confusion.} A column of a confusion matrix charts the distribution of predictions of a classifier for data in a particular category. We call such a distribution the \emph{classification distribution}. We adapt this notion for our first collision detection approach, which identifies a candidate intent $A$ to collide with $B \in \mathcal{C}$ if a classifier model trained on dataset $\mathcal{C}$ produces a \emph{classification distribution} $d$ such that $\frac{max(d)}{sum(d)} > \tau$, where $\tau$ is a threshold set by the developer. We call this ratio the \emph{classifier collision score}.


\begin{table}[]
    \centering
    \scalebox{0.9}{
    \begin{tabular}{lrr}
    \toprule
        \textbf{Dataset} & \textbf{\# Intents} & \textbf{\# Collisions} \\
        \midrule
        \emph{ACID} & 175 & 36\\
        \emph{Clinc-150} & 150 & 158\\
        \emph{MTOP} & 113 & 60\\
        \emph{Banking-77} & 77 & 25\\
        \emph{HWU} & 64 & 103\\
        \emph{New} & 58 & 5\\
        \emph{MetalWOz} & 51 & 80\\
        \emph{DSTC-8} & 34 & 67\\
        \emph{ATIS} & 26 & 7\\
        \emph{Outlier} & 10 & 9\\
        \emph{Snips} & 7 & 20\\
        \emph{Jobs640} & 1 & 0\\
        \emph{Talk2Car} & 1 & 0\\
        \midrule
        Total & 767 & 570\\
        \bottomrule
    \end{tabular}}
    \caption{Number of intents with collisions. A total of 570 intents have at least one collision.}
    \label{tab:collision_statistics}
\end{table}

\paragraph{Data Coverage.} We define the \emph{coverage} of one intent $B$ over another intent $A$ as 
$$
    \text{Coverage}(A, B) = \frac{1}{|B|} \sum_{b\in B} \max_{a \in A} sim(a, b).
$$
\noindent Here, $sim(a,b)$ computes the similarity between two phrases $a$ and $b$ (for instance, $sim(a,b)$ could be the cosine similarity between two phrase embeddings or the Jaccard similarity between n-\emph{gram} sets). The coverage metric can be used to detect if two intents collide using a threshold rule. In other words, $A$ and $B$ collide if Coverage($A$,$B$) $> \kappa$, where $\kappa$ is a threshold chosen by the developer.
We call the coverage metric the \emph{coverage score}.




\begin{figure}
    \centering
    \includegraphics[clip,width=\columnwidth]{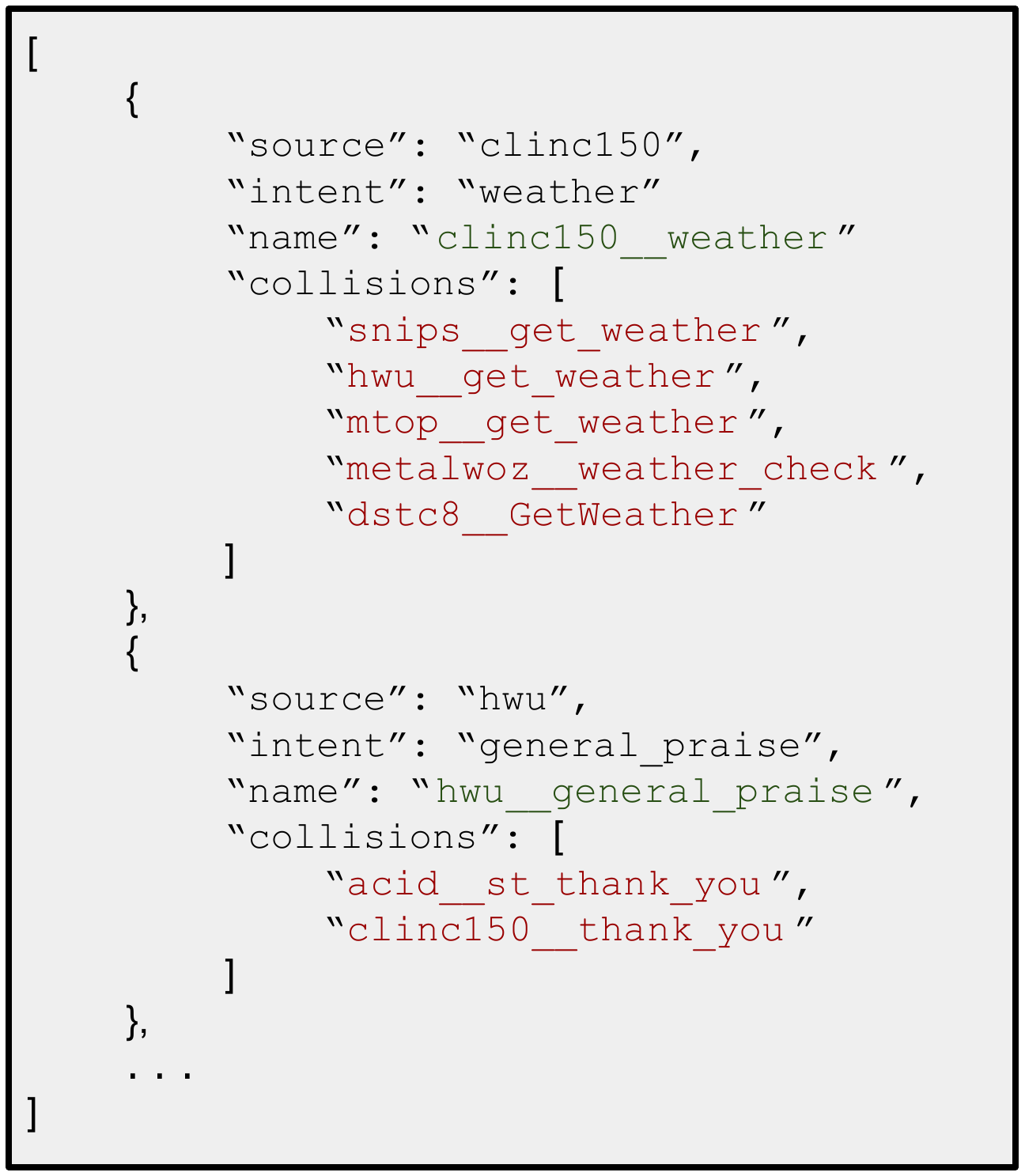}
    \caption{Example entries in the graph-like collision meta-dataset, showing collisions for \emph{Clinc-150}'s \texttt{weather} intent and \emph{HWU}'s \texttt{general\_praise} intent.}
    \label{fig:metadata_json_listing}
\end{figure}

\subsection{Datasets}\label{sec:datasets} 

We evaluate the effectiveness of our intent collision approaches using several indicative datasets.
These datasets can be roughly grouped into three categories:
(1) intent classification datasets like  \emph{Clinc-150} \cite{chatbot150}, \emph{Banking-77} \cite{banking77}, \emph{ACID} \cite{ACID2020}, \emph{Outlier} \cite{outlier}, and \emph{New} (this work; a corpus that was crowdsourced in a manner similar to \citet{chatbot150} and \citet{outlier}); (2) joint slot-filling and intent classification or semantic parsing datasets like \emph{ATIS} \cite{hemphill-etal-1990-atis, hirschman-1992-multi-location-atis, hirschman-etal-1993-multi-lication-atis, dahl-etal-1994-expanding-atis}, \emph{Snips} \cite{snips}, \emph{HWU} \cite{sds-eval}, and \emph{MTOP} \cite{li2021mtop}; and (3) turn-based dialog datasets like \emph{DSTC-8} \cite{dstc8} and \emph{MetalWOz} \cite{MetalWOz}.
We only consider the initial queries in the turn-based \emph{DSTC-8} and \emph{MetalWOz}, and discard all subsequent dialog turns.

Queries in these datasets span a wide range of topic domains, including banking and personal finance (\emph{Banking-77} and \emph{Outlier}) and insurance (\emph{ACID});
other datasets cover a wide array of topic domains, such as \emph{Clinc-150} and \emph{HWU}, which cover smart home, automotive, travel, banking, cooking, and others. 
Since we are concerned with detecting colliding intents, we do not consider any slot annotations, and we use only the first turns from the multi-turn dialog datasets.
In addition, we also use the \emph{Jobs640}~\cite{jobs640} and \emph{Talk2Car}~\cite{talk2car} datasets, which, although not originally designed for intent classification tasks, are categorized in a way that admit consideration as single-intent classification for our purposes.
Table~\ref{tab:collision_statistics} summarizes these datasets.

\paragraph{The Collision Meta-Dataset}

We constructed a graph-like dataset that indicates the collision relationships between intents.
To build this dataset, we reviewed all intents from all of the datasets listed in Table~\ref{tab:collision_statistics} to check for collisions between other intents.
We developed a ground truth set of tuples indicating whether two intents collide among these datasets.
Figure~\ref{fig:metadata_json_listing} shows the structure of the intent collision meta-dataset, and Table~\ref{tab:collision_statistics} displays the number of collisions that occur relative to each individual dataset.
The meta-dataset includes the three types of collisions defined in Section~\ref{sec:problem-statement}.

\subsection{Experimental Evaluation}

\paragraph{Implementation Details.}

We evaluate our intent collision detection methods on our newly-created collision meta-dataset.
For evaluating the classifier confusion approach, we train a multi-class intent classifier on each individual dataset (except the single-intent datasets) and then run inference on all other intents from the other datasets.
We compute and report the classifier confusion score for each run.
In our experiments, we use a linear SVM classifier with bag-of-words feature representations.

For evaluating the data coverage approach, we first sample\footnote{Sampling avoids combinatorial explosion of possible intent pairs.} a nearly equal number of colliding and non-colliding intent pairs from the collision meta-dataset.
We then compute the coverage scores for the selected pairs using several sentence representation and similarity metrics.
We use the SBERT library's SBERT-NLI and SBERT-miniLM sentence embedders~\cite{reimers-gurevych-2019-sentence-bert} along with cosine similarity.
Additionally, we also use \emph{n}-gram-based similarity, defined as 
$$
sim(a,b) = \frac{1}{N}\sum_{n=1}^{N} \frac{|\text{\emph{n}-grams}_a \cap \text{\emph{n}-grams}_b|}{|\text{\emph{n}-grams}_a \cup \text{\emph{n}-grams}_b|}
$$
where \emph{a} and \emph{b} are queries from two intents, and $N=3$ in our experiments.

For both the data coverage and classifier confusion experiments, we only consider intents that have at least 10 queries. 
For the collision detection experiments, we used all 285 collision pairs and sampled 300 non-colliding pairs since there are substantially more non-colliding pairs.
The classifier confusion approach does not compare intents in a pairwise manner, and instead compares a dataset (i.e., a classifier trained on a dataset) against a single intent at a time.
We run a classifier on all multi-intent datasets,
which yielded a total of 400 collision pairs and 6,802 non-collision pairs for the classifier confusion experiments. 

\begin{figure}[t]

\subfloat[SBERT-NLI Coverage Score]{%
  \includegraphics[clip,width=\columnwidth]{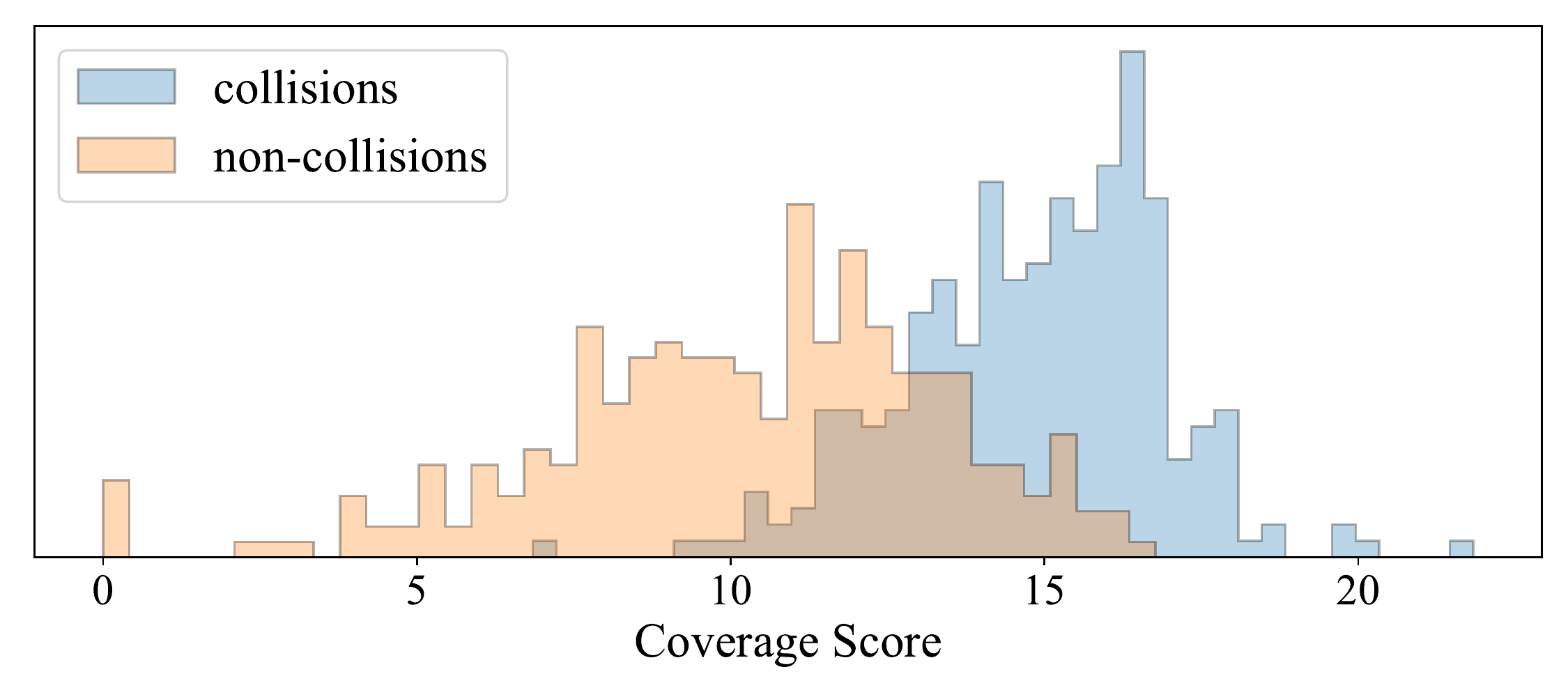}%
}

\subfloat[Mini-LM Coverage Score]{%
  \includegraphics[clip,width=\columnwidth]{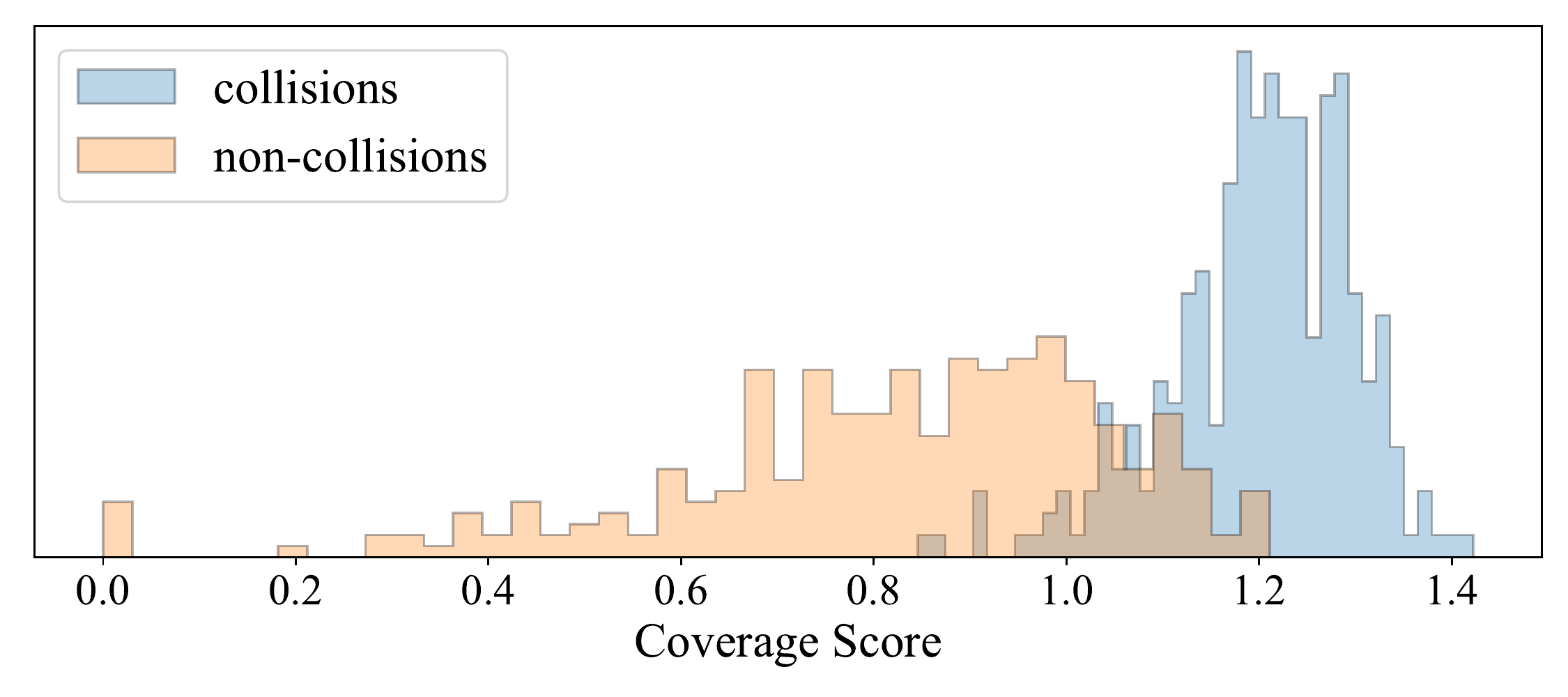}%
}

\subfloat[SVM classifier confusion.]{%
  \includegraphics[clip,width=\columnwidth]{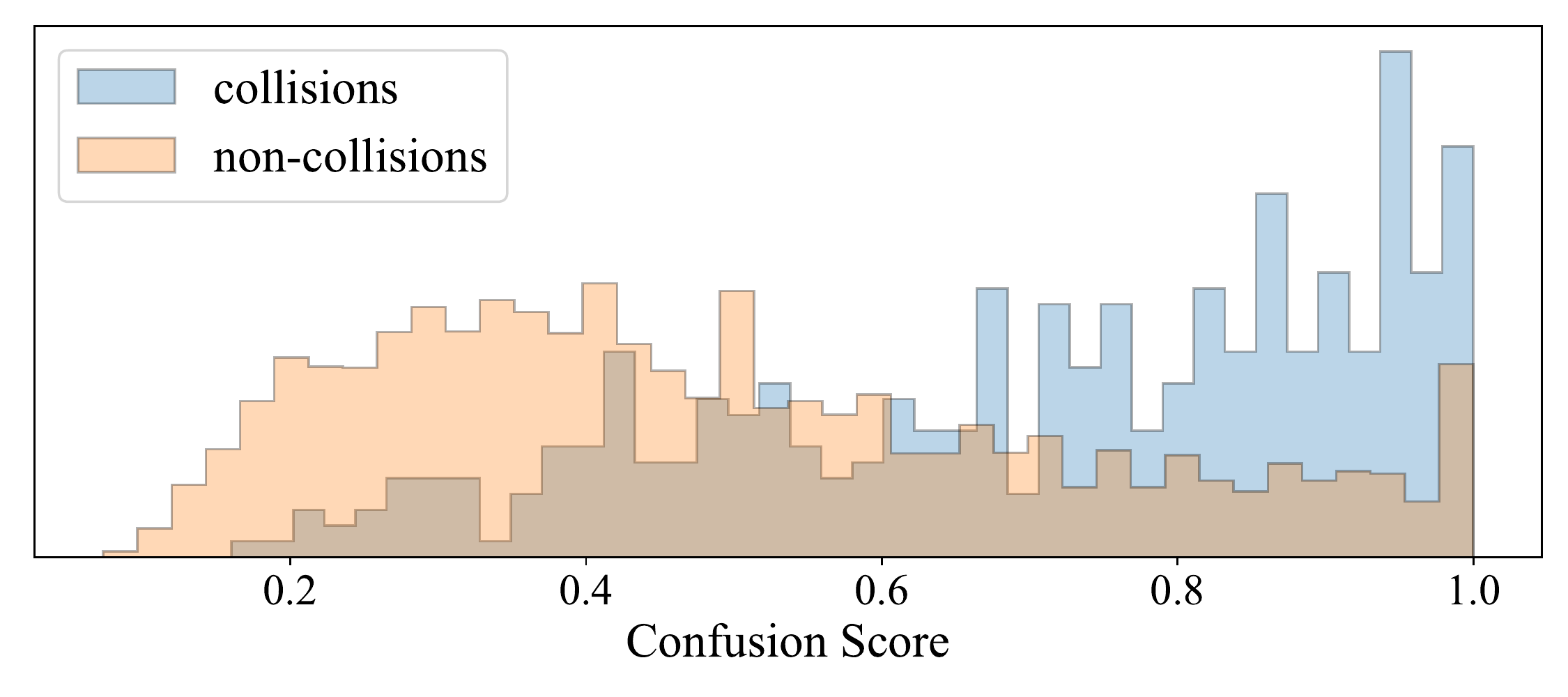}%
}

\caption{Data coverage and classifier confusion score distributions for various intent collision detection approaches.}
\label{fig:collision-score-distributions}
\end{figure}

\paragraph{Metrics.}
While in actual application settings, a user may wish to use thresholds for $\tau$ and $\kappa$ (defined earlier in Section~\ref{sec:collision-detection-methods}) to determine whether intents collide, we evaluate both classifier confusion and coverage methods in a threshold-free manner using the AUC score. (In practice, values for $\tau$ and $\kappa$ could be set by the practitioner via cross-validation or by using the meta-dataset provided in this work to set optimal thresholds for their application.)
The AUC score allows us to judge each method's ability to distinguish collisions versus non-collisions;
an AUC score of 1.0 means perfect separability between collisions and non-collisions, while an AUC score of 0.5 means a method is unable to distinguish between colliding and non-colliding intents. 




\subsection{Results}
\paragraph{Data Coverage.}

Figure~\ref{fig:collision-score-distributions} charts coverage scores and confusion scores for various approaches.
In Figure~\ref{fig:collision-score-distributions} (a) and (b), the coverage approaches tend to return higher coverage scores for non-collisions and lower coverage scores for collisions, which aligns with our expectations given our definition of the coverage metric and assuming the similarity metric used in the coverage computation is effective.
The AUC scores allow us to quantitatively judge the performance of the various coverage-based approaches: in Table~\ref{tab:collision-detection-reslts}, the SBERT-miniLM embedding method yields the highest AUC score, and interestingly the \emph{n}-gram-based coverage method performs second best, with the SBERT-NLI embedding method in third.

\begin{table}[]
    \centering\scalebox{0.85}{
    \begin{tabular}{lcc}
    \toprule
       & \textbf{Coverage} & \textbf{Confusion} \\
        \textbf{Approach}  & \textbf{AUC} & \textbf{AUC}  \\
    \midrule
        SBERT-NLI & 0.898 & --- \\
        token & 0.931 & --- \\
        SBERT-miniLM & 0.963 & --- \\
        SVM-based & --- & 0.756\\
        \bottomrule
    \end{tabular}}
    \caption{AUC metrics for each intent collision detection approach.}
    \label{tab:collision-detection-reslts}
\end{table}

\begin{table*}[t]
    \centering\scalebox{0.78}{
    \begin{tabular}{l l l}
    \toprule
        \textbf{Original} & \textbf{Original} & \\
        \textbf{Dataset} & \textbf{Intent} & \textbf{Sample}  \\
    \midrule
         \emph{HWU} & \texttt{alarm\_remove} &  \emph{remove the alarm set for 10pm}\\
         \emph{Clinc-150}& \texttt{reminder\_update} & \emph{set a reminder for me to take my meds} \\
        \emph{MTOP} & \texttt{get\_weather} &  \emph{should i wear a raincoat tuesday} \\
        \emph{Jobs-640} & ---  & \emph{what systems analyst jobs are there in austin}\\
        \emph{Talk2Car} & ---  & \emph{switch to right lane and park on right behind parked black car} \\
        \emph{Jobs640} & \texttt{Jobs640} & \emph{what systems analyst jobs are there in austin} \\
        \emph{Snips} & \texttt{add\_to\_playlist} & \emph{add paulinho da viola to my radio rock song list} \\
        \emph{Outlier} & \texttt{hours} & \emph{tell me the hours of operation for my bank} \\
        \emph{New} & \texttt{balance} & \emph{do I have holiday time saved} \\
        \emph{DSTC-8} & \texttt{LookupMusic} & \emph{I like metal songs can you find me some} \\
        \emph{ATIS} & \texttt{ground\_service} & \emph{i'll need to rent a car in washington dc} \\
        \emph{MetalWOz} & \texttt{name\_suggester} & \emph{I need to find a name for my new cat} \\
        \emph{Clinc-150} & \texttt{find\_phone} & \emph{can you help me find my cell} \\
        \emph{ACID} & \texttt{info\_amt\_due} & \emph{what is the current amount due on my account} \\
        \emph{Banking-77} & \texttt{terminate\_account} & \emph{how do I deactivate my account} \\
        \emph{Clinc-150} & \texttt{measurement\_conversion} & \emph{what amount of millimeters are in 50 kilometers} \\
        \emph{ACID} & \texttt{info\_name\_change} & \emph{i need to fix my name} \\
        \emph{MTOP} & \texttt{play\_music} & \emph{find me the latest linkin park album} \\
        \emph{HWU} & \texttt{audio\_volume\_up} & \emph{just increase the volume a little} \\
        \emph{Outlier} & \texttt{balance} & \emph{how much oney do i have available}\\
    \midrule
     \citet{vertanen_aacdialogue} & --- & \emph{why on earth is there cereal in the fridge}\\
     \citet{vertanen_aacdialogue} & --- & \emph{who are you going to vote for in november} \\
     \citet{vertanen_aacdialogue} & --- & \emph{do you know where i put my glasses}\\
     \emph{Clinc-150} & \texttt{out-of-scope} & \emph{what size wipers does this car take} \\
     \emph{Clinc-150} & \texttt{out-of-scope} & \emph{how long is winter} \\
     \emph{Clinc-150} & \texttt{out-of-scope} & \emph{are any earning reports due} \\
    \bottomrule
    \end{tabular}}
    \caption{Sample intents and queries from our \emph{Redwood} dataset, along with the corresponding original dataset and intent (where applicable). Samples are grouped into \emph{in-scope} (top) and \emph{out-of-scope} (bottom).}
    \label{tab:redwood-examples}
\end{table*}

\paragraph{Classifier Confusion.}

Figure~\ref{fig:collision-score-distributions} (c) charts classifier confusion scores for the SVM-based classifier confusion approach.
Our results demonstrate that actual intent collisions typically yield high classifier confusion scores, while non-collisions yield lower confusion scores.
Visually, however, Figure~\ref{fig:collision-score-distributions} (c) seems to indicate that that the classifier confusion approach is less effective than the coverage-based approaches.
This is made more apparent by the AUC score in 
Table~\ref{tab:collision-detection-reslts}.
We note that the data coverage and classifier confusion AUC scores are not directly comparable as they use different evaluation settings. Nonetheless, the difference in performance scores does lead us to conclude that the data coverage approach is more effective.

In sum, these experimental results demonstrate that the two intent collision detection approaches introduced here are effective in detecting collisions among real datasets, with the data coverage approach being the stronger of the two.

\section{Building the \emph{Redwood} Dataset}

With tools addressing the problem of intent collision detection in hand, we now turn our attention to combining the individual datasets from Table~\ref{tab:collision_statistics} together to form a single large-scale intent classification dataset, \emph{Redwood}.
This section discusses the construction of \emph{Redwood} and a companion \emph{out-of-scope} evaluation set, and then evaluates several benchmark intent classifiers on the dataset. 
These datasets and associated evaluations demonstrate the consequences of leaving colliding intents unaddressed, providing a valuable resource for the community to improving intent classification models.

\subsection{Data}

\paragraph{\emph{In-Scope} Data.} After creating the collision meta-dataset, a natural extension was to combine each dataset together to form \emph{Redwood}.
We used the collision meta-dataset to help inform us of which intents could combined, and which intents could stand alone in \emph{Redwood}. 
In some cases, we removed intents that caused \emph{hierarchical} collisions, as sometimes joining together intents from a hierarchical collision produced an intent that was too broad. 
We included only those intents that have at least 50 queries, and the resulting \emph{Redwood} consists of \RedwoodSize~total intents and \RedwoodQueries~queries.
Following the terminology used in \citet{chatbot150}, we call these \RedwoodSize~intents \emph{in-scope}.

\begin{table}[]
    \centering\scalebox{0.85}{
    \begin{tabular}{lc}
    \toprule
        \textbf{Dataset} & \textbf{N. Samples}  \\
        \midrule
        \citet{vertanen_aacdialogue} & 2067 \\
        \emph{Clinc-150} & 1200\\
        \midrule
        Total & 3267\\
        \bottomrule
    \end{tabular}}
    \caption{Sources of \emph{out-of-scope} data and number of samples used in \emph{Redwood}'s out-of-scope test set.}
    \label{tab:my_label}
\end{table}

By way of comparison, we also produced a "na\"{i}ve" version of \emph{Redwood}, called \emph{Redwood-na\"{i}ve}, where all the intents from the datasets listed in Table~\ref{sec:datasets} were joined together \emph{without} using collision detection or any other method of arbitrating or correcting colliding intents.
Like the original \emph{Redwood}, we included only intents that have at least 50 queries, and capped each intent at a maximum of 150 queries so as to avoid drastic class imbalances.
\emph{Redwood-na\"{i}ve} consists of \RedwoodNaiveSize~intents and \RedwoodNaiveQueries~total queries.

All versions of \emph{Redwood} were split into train and test splits per intent: 85\% training, 15\% testing. 


\paragraph{\emph{Out-of-Scope} Data.} In contrast to \emph{in-scope}, \emph{out-of-scope} queries are those that do not belong to any of the \emph{in-scope} intents.
Considering \emph{out-of-scope} queries in an evaluation of intent classification models is important because such queries occur in production settings, where end users cannot be expected to know the full range of intents when interacting with a conversational AI system.
We include a collection of \OOSTotalQueries~\emph{out-of-scope} queries in addition to the \emph{Redwood} corpus.
\emph{Redwood}'s \emph{out-of-scope} data originates from the following sources: \emph{Clinc-150} dataset, which itself includes a set of \emph{out-of-scope} queries; and \citet{vertanen_aacdialogue}, a crowdsourced dialog dataset from which we use the first dialog turns.
We reviewed all candidate \emph{out-of-scope} queries, removing those that were actually \emph{in-scope}.
Examples of queries from the \emph{Redwood} dataset are shown in Table~\ref{tab:redwood-examples}.

\subsection{Benchmark Evaluation}

\paragraph{Models.}
We benchmark intent classification performance using the MobileBERT model \cite{sun-etal-2020-mobilebert} using the HuggingFace library \cite{wolf-etal-2020-transformers-huggingface}.
The MobileBERT implementation uses a softmax function to compute logits to a probability vector \textbf{\emph{p}}, from which we can obtain confidence scores for each intent. These confidence scores can be used to predict whether a query is in- or out-of-scope, according to a decision threshold \emph{t} given by 
  \begin{equation*}
    \text{decision rule}=
    \begin{cases}
      \text{in-scope}, & \text{if}\ \text{max}(\emph{\textbf{p}}) \ge t \\
      \text{out-of-scope}, & \text{if}\ \text{max}(\emph{\textbf{p}}) < t.
    \end{cases}
  \end{equation*}
Such decision rules were used in \citet{hendrycks17baseline} and \citet{chatbot150}.

\paragraph{Metrics and Experiments.}

We measure intent classifier accuracy on in-scope data without considering out-of-scope inputs.
We also measure each model's ability to distinguish in-scope and out-of-scope queries by computing the AUC between in- and out-of-scope confidence scores.
In this way, we use AUC to measure how separable in- and out-of-scope queries based on their confidence scores without having to select an confidence threshold \emph{t}.
An AUC score of 0.5 (the minimum AUC score) implies the model cannot distinguish in- versus out-of-scope inputs.
An AUC of 1.0 indicates the model can perfectly separate inputs.


\subsection{Results}

\begin{table}[]
    \centering
    \scalebox{0.84}{
    \begin{tabular}{lccc}
    \toprule
        \textbf{Training} & \textbf{In-Scope} & \textbf{Clinc} & \textbf{Vertanen}\\
        \textbf{Dataset} & \textbf{Accuracy} & \textbf{OOS AUC} & \textbf{OOS AUC}\\
    \midrule
        \emph{Redwood} & 0.913 & 0.921 & 0.928\\
        \emph{Redwood-na\"{i}ve} & 0.861 & 0.909 & 0.925\\
    \bottomrule
    \end{tabular}}
    \caption{Model performance of the MobileBERT classifier on \emph{Redwood} and \emph{Redwood-na\"{i}ve}.}
    \label{tab:redwood-classifier-results}
\end{table}

\begin{table}[]
    \centering\scalebox{0.7}{
    \begin{tabular}{lcccccccc}
    \toprule
        \textbf{Collisions} & 0 & 1 & 2 & 3 & 4 & 5 & 6 & 14 \\
    \midrule
       Mean Acc. & 0.91 & 0.80 & 0.81 & 0.79 & 0.81 & 0.80 & 0.89 & 0.57 \\
       Size & 322 & 74 & 42 & 51 & 15 & 11 & 13 & 8 \\
    \bottomrule
    \end{tabular}}
    \caption{Accuracy scores on \emph{Redwood-na\"{i}ve} intents per number of collisions.}
    \label{tab:naive_table}
\end{table}

Model performance on \emph{Redwood-na\"{i}ve} and \emph{Redwood} is shown in Table~\ref{tab:redwood-classifier-results}.
First, we notice that the intent classifiers perform reasonably well on the in-scope classification task, with MobileBERT classifying queries with 91\% accuracy.
The models also perform well on the out-of-scope task, and discriminate between in- and out-of-scope queries with AUC scores of 0.921 and 0.928 on the \emph{Clinc-150} and \citet{vertanen_aacdialogue} out-of-scope data.

The bottom half of Table~\ref{tab:redwood-classifier-results} presents model performance when trained and tested on \emph{Redwood-na\"{i}ve}.
In this case, model performance is substantially worse than models trained on the carefully-crafted \emph{Redwood} dataset, confirming our hypothesis from Section~\ref{sec:motivating-example} that model performance suffers if trained on data with colliding intents.

We drill deeper into the impact of intent collisions on models trained on \emph{Redwood-na\"{i}ve} in
Table~\ref{tab:naive_table}
which charts per-intent accuracy based on the number of other intents that collide with that intent.
This table groups intents based on the number of collisions, and we see that on average, intents with no collisions exhibit higher accuracy than intents with collisions.
In general, colliding intents lead to degraded accuracy: intents with one or more collisions have accuracy of around 10 or more points lower than the no-collision group, with the exception of the 6-collision group.
The average accuracy of the 6-collision group on \emph{Redwood-na\"{i}ve} is indeed surprising, and we posit that the MobileBERT model---a high-capacity transformer model---can learn the nuances of each individual intent, even if they do semantically collide.


\section{Conclusion and Future Work}

This paper introduces the task of intent collision detection when constructing or updating an intent classification model's dataset to incorporate additional intents.
Using \RedwoodDatasets~individual datasets, we constructed a meta-dataset to track intent collisions between the datasets,
and then introduced and evaluated two intent collision detection techniques and found that both perform effectively at the collision detection task.
To help measure and address this problem, we constructed \emph{Redwood}, a large-scale intent classification dataset consisting of \RedwoodSize~intents and over 60,000 queries.
We used \emph{Redwood} to benchmark several intent classification models on the task of in-scope query prediction and out-of-scope detection,
The new \emph{Redwood} dataset is the largest publicly available intent classification benchmark, in terms of number of intents,
and will be made publicly available.
Future work will include annotating slots to extend Redwood to joint intent classification and slot-filling, and it is likely that new tools will have to be developed for doing so.
Additionally, using the collision detection methods introduced in this paper, \emph{Redwood} can be periodically updated with new intents whenever other new intent classification datasets are published.

\section*{Acknowledgements}
We thank the anonymous reviewers for their detailed and thoughtful feedback, and Jacob Solawetz for his feedback on early iterations of the \emph{Redwood} concept.

\bibliography{acl_latex}



\end{document}